\pdfoutput=1
\documentclass[11pt]{article}

\usepackage[final]{ACL2023}

\usepackage{times}
\usepackage{latexsym}
\usepackage{url}
\usepackage{arydshln}
\usepackage{enumitem}
\usepackage{float}
\usepackage{amsmath}
\usepackage{amsfonts,bm}
\makeatletter   
\newcommand{\sveryshortarrow}[1][3pt]{\mathrel{%
    \vcenter{\hbox{\rule[-.5\fontdimen8\scriptfont3]
               {\scriptratio\dimexpr#1\relax}{\fontdimen8\scriptfont3}}}%
   \mkern-4mu\hbox{\let\f@size\sf@size\usefont{U}{lasy}{m}{n}\symbol{41}}}}
\makeatother

\def\eqref#1{equation~\ref{#1}}
\def\1{\bm{1}}

\def\m1{{\bm{1}}}

\DeclareMathAlphabet{\mathsfit}{\encodingdefault}{\sfdefault}{m}{sl}
\SetMathAlphabet{\mathsfit}{bold}{\encodingdefault}{\sfdefault}{bx}{n}

\usepackage{xspace}

\usepackage{amssymb}
\usepackage{amsmath}
\usepackage{tikz,pgfplots}
\usepackage{algorithm}
\usepackage{algorithmic}
\renewcommand{\algorithmiccomment}[1]{\bgroup\hfill//~#1\egroup}
\usepackage[switch]{lineno}  %
\usepackage{caption}
\usepackage{subcaption}
\usepackage{booktabs, multirow}
\usepackage{comment}
\usepackage{balance}
\usepackage{cleveref}
\crefname{section}{§}{§§}
\Crefname{section}{§}{§§}

\usepackage[T1]{fontenc}
\usepackage[utf8]{inputenc}

\usepackage{microtype}

\usepackage{inconsolata}
\usepackage{pgfplots}
\usepackage{tikz}
\usepackage{mathdots}
\pgfplotsset{compat=1.14}
\definecolor{bblue}{HTML}{4F81BD}
\definecolor{rred}{HTML}{C0504D}
\definecolor{ggreen}{HTML}{9BBB59}
\definecolor{ppurple}{HTML}{9F4C7C}
\pgfplotsset{
  compat=1.14,
  legend entry/.initial=,
  every axis plot post/.code={%
      \pgfkeysgetvalue{/pgfplots/legend entry}\tempValue
      \ifx\tempValue\empty
          \pgfkeysalso{/pgfplots/forget plot}%
      \else
          \expandafter\addlegendentry\expandafter{\tempValue}%
      \fi
  },
}
\title{Mimic-IV-ICD: A new benchmark for eXtreme MultiLabel Classification}

\newcommand*{\affaddr}[1]{#1} % No op here. Customize it for different styles.
\newcommand*{\affmark}[1][*]{\textsuperscript{#1}}
\newcommand*{\email}[1]{\textrm{#1}}
\author{
Thanh-Tung Nguyen\affmark[1]~~
Viktor Schlegel\affmark[1,3]~~
Abhinav Kashyap\affmark[1]~~
Stefan Winkler\affmark[1,2]~~ \\
\textbf{Shao-Syuan Huang\affmark[1,4]}~~ 
\textbf{Jie-Jyun Liu\affmark[4]}~~
\textbf{Chih-Jen Lin\affmark[4,5]}~~\\
\affaddr{\affmark[1]AICS Singapore / TaiPei} \\
\affaddr{\affmark[2]National University of Singapore} \\
\affaddr{\affmark[3]University of Manchester, United Kingdom} \\
\affaddr{\affmark[4]National Taiwan University} \\
\affaddr{\affmark[5]Mohamed bin Zayed University of Artificial Intelligence}\\
\email{\small{\{thomas\_nguyen; viktor\_schlegel; abhinav\_kashyap; stefan\_winkler;gordon\_huang\}@asus.com}}\\
\email{\small{jiejyun02@gmail.com; cjlin@csie.ntu.edu.tw}}\\
}

\begin{document}
\maketitle
\begin{abstract}
Clinical notes are assigned ICD codes -- sets of codes for diagnoses and procedures. In the recent years, predictive machine learning models have been built for automatic ICD coding. However, there is a lack of widely accepted benchmarks for automated ICD coding models based on large-scale public EHR data.

This paper proposes a public benchmark suite for ICD-10 coding using a large EHR dataset derived from MIMIC-IV, the most recent public EHR dataset. We implement and compare several popular methods for ICD coding prediction tasks to standardize data preprocessing and establish a comprehensive ICD coding benchmark dataset. This approach fosters reproducibility and model comparison, accelerating progress toward employing automated ICD coding in future studies. Furthermore, we create a new ICD-9 benchmark using MIMIC-IV data, providing more data points and a higher number of ICD codes than MIMIC-III. Our open-source code offers easy access to data processing steps, benchmark creation, and experiment replication for those with MIMIC-IV access, providing insights, guidance, and protocols to efficiently develop ICD coding models.
\end{abstract}

\section{Introduction}
Medical records and clinical documentation serve as essential sources of information on patient care, disease progression, and healthcare operations. Following a patient's visit, medical coders analyze these records and identify diagnoses and procedures according to the International Classification of Diseases (ICD) system \citep{ICD}. These codes are used in predictive modelling for patient care and health status, as well as for insurance claims, billing processes, and other hospital functions \citep{Tsui2002ValueOI}.

\begin{figure*}[t!]
\centering
\resizebox{\textwidth}{!}{\includegraphics[clip, trim=0.1cm 0.1cm 0.7cm 1cm, width=1.0\textwidth]{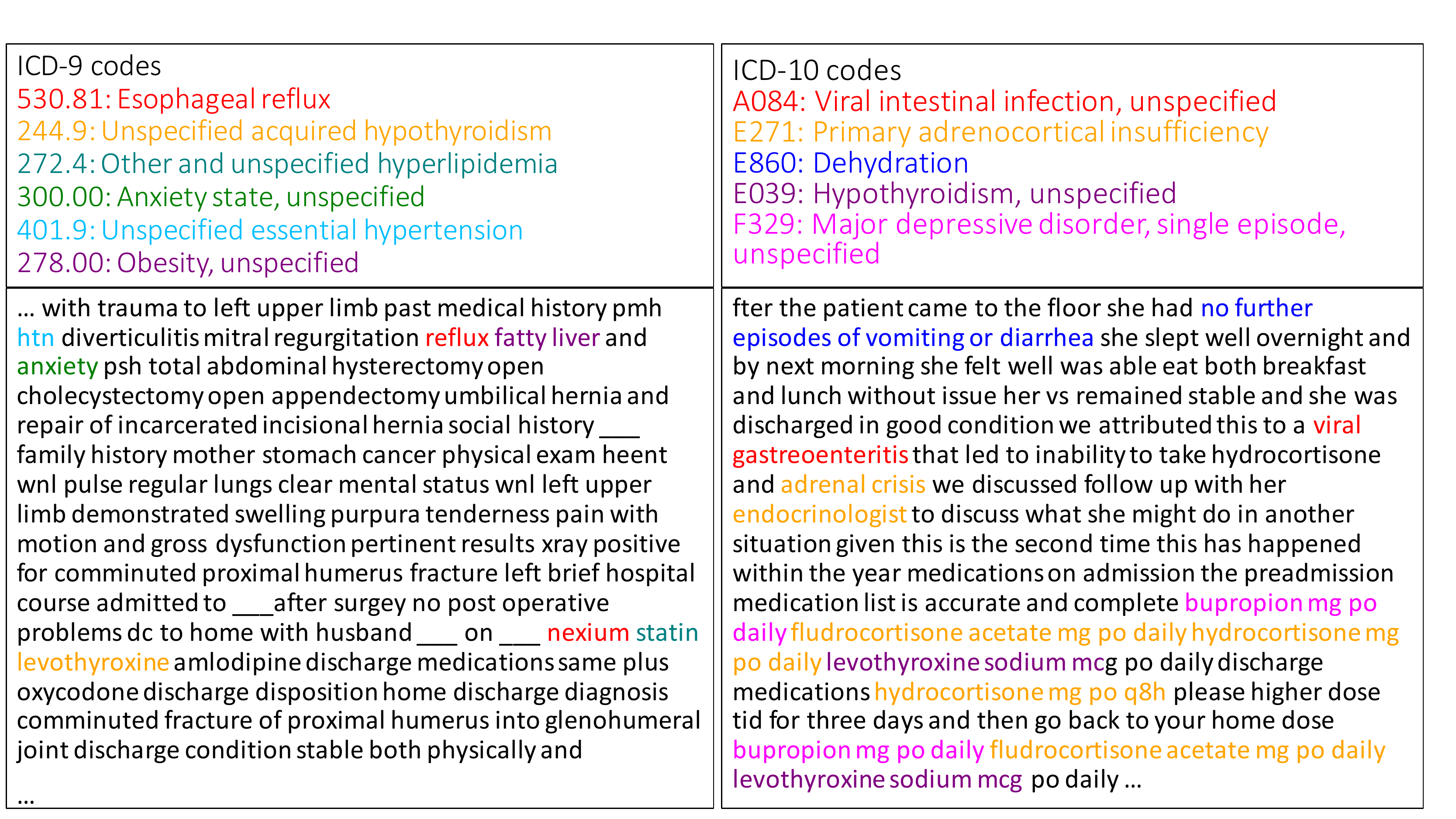}}%{imgs/MIMIC-IV-ICD_ver2.png}
% \vspace{-1em}
\caption{\small Two example EHR notes from the MIMIC-IV dataset, featuring both ICD-9 and ICD-10 codes (top) alongside their corresponding discharge note text (bottom). For clarity, each code and its related mentions or evidence within the note text are color-coded. }
% \red{make it single column. we may need more space}}
\label{fig:Examples of EHR note from the MIMIC-IV}
\end{figure*}

Despite significant progress in healthcare, manual processes continue to present challenges. One such challenge is manual ICD coding, which requires interpreting complex medical records with an extensive vocabulary and limited content. Coders must select a small subset from a continuously expanding list of ICD codes (18,000 codes in ICD-9 and 155,000 codes in ICD-10). Figure~\ref{fig:Examples of EHR note from the MIMIC-IV} displays discharge summaries from the MIMIC-IV data set together with their assigned ICD-9 and ICD-10 codes. 
For each potential code, the coder must examine the entire medical note to identify phrases relevant to the code description. For instance, "viral gastroenteritis," a common stomach and intestinal infection, is related to code \emph{A00.84}, ``Viral intestinal infection''. Scanning through thousands of tokens in the medical note text to match with thousands of code descriptions increases the likelihood of coding errors. Such errors potentially lead to financial losses or misallocation of resources in patient care. Consequently, there is a growing interest in automated ICD coding within both industry and academia.

Before the advent of deep learning techniques, automated ICD coding methods relied on rule-based or decision tree-based approaches \citep{rulebaseicd2008,scheurwegs2017selecting}. The focus has since changed to neural networks, which can be classified into two main categories. The first involves encoding medical documents using pre-trained language models \citep{li2020multirescnn, liu-etal-2021-effective}, adapting pre-trained language models for the clinical domain \citep{lewis-etal-2020-pretrained}, or improving language models with medical knowledge, such as disease taxonomies, synonyms and abbreviations \citep{Yang2022KnowledgeIP,yuan-etal-2022-code}. The second category aims to improve pre-trained language model representations by capturing the relevance between the document and label metadata, including descriptions \citep{mullenbach-etal-2018-explainable,ijcai2020-461-vu}, co-occurrences \citep{cao-etal-2020-hypercore}, hierarchies \citep{falis-etal-2019-ontological,ijcai2020-461-vu}, or thesaurus knowledge, such as synonyms \citep{yuan-etal-2022-code}.

Although these approaches are intended to address challenges specific to medical coding, such as specialized vocabulary and a large number of labels, their main limitation is their reliance on a single dataset---MIMIC-III \citep{johnson2016mimic}---and model settings tailored specifically to that dataset. The medical notes in MIMIC-III cover only around 9,000 codes, whereas nearly twice that number is used in practice. Furthermore, there is a lack of comparative studies among different methods and models under more extreme conditions with longer-tailed distributions and significantly higher numbers of ICD codes. % More importantly, the practical value of an ICD-9 benchmark such as MIMIC-III is questionable, as the ICD-9 coding system has been phased out in favor of ICD-10. 
Using a comprehensive public EHR database, our objective was to standardize data preprocessing and establish an extensive ICD coding benchmark data set to facilitate reproducibility and model comparison, ultimately accelerating progress toward incorporating automated ICD coding into future studies.

The extensive adoption of Electronic Health Records (EHR) has resulted in the accumulation of vast amounts of data that can be used to develop predictive models to improve ICD coding. Using EHR databases, such as the Medical Information Mart for Intensive Care III (MIMIC-III), several benchmarks have been established for ICD-9 coding in full-code and high-frequent code settings \citep{mullenbach-etal-2018-explainable, shi-et-al-2027-automated-icd}. These benchmarks have standardized the conversion of raw medical notes into data suitable for building predictive models. They offer clinicians and researchers easy access to high-quality medical data, accelerating research and validation efforts \citep{ijcai2020-461-vu}. Non-proprietary databases and open-source pipelines enable the reproduction and enhancement of clinical studies in previously unattainable ways. Since its publication in 1977, ICD-9 contains outdated and obsolete terms that are no longer relevant to current medical practices. The structure of ICD-9 limits the number of new codes that can be created and added to the code list, and many ICD-9 categories have reached their capacity. In contrast, ICD-10, which was introduced in 1992, has been designed to accommodate code expansion, allowing healthcare providers to use codes that are more specific to patient diagnoses. Although there are some publicly available benchmarks, the majority focus on the coding settings of ICD-9, with no widely recognized benchmarks for ICD-10. A public ICD-10 benchmark would reduce entry barriers for new researchers and facilitate new research.

\begin{table*}[ht]
    \centering
    \small
    \begin{tabular}{lccc}
    \toprule
    & MIMIC-III Full & MIMIC-IV-ICD9 & MIMIC-IV-ICD10  \\
    \midrule
    \# Doc. & 52,726 & 209,359 & 122,317 \\
    Avg \# of words per Doc. & 1,462 & 1,460 & 1,662 \\
    Avg \# of ICD codes per Doc. & 13.9 & 13.4 & 16.1 \\ 
    Total \# of Unique ICD codes & 8,922 & 11,331 & 26,096 \\        
    \bottomrule
    
    \end{tabular}
    \caption{Statistics of master datasets of MIMIC-III and MIMIC-IV under ICD-9 and ICD-10 codes settings.}
    \label{tab: statistics-master-dataset}
\end{table*}

In this paper, we propose a public benchmark suite for ICD-10 coding using a large data set derived from MIMIC-IV, the most recent public EHR data set containing a decade of critical care database from 2008 to 2019 with 431,231 hospital admissions and 180,733 unique patients. We implement and compare several popular methods for the ICD coding prediction tasks. In addition to ICD-10 coding, we used MIMIC-IV data to create a new ICD-9 benchmark with more data points and a greater number of ICD codes than MIMIC-III. Our open source code \footnote{https://github.com/thomasnguyen92/MIMIC-IV-ICD-data-processing} allows anyone with access to MIMIC-IV to follow our data processing steps, generate benchmarks, and reproduce our experiments. This study equips future researchers with information, recommendations, and protocols for processing raw data and developing ICD coding models efficiently and promptly.

\section{ICD Code Benchmark}
\subsection{Data Processing}

\begin{figure}[t!]
\centering
\resizebox{\columnwidth}{!}{\includegraphics[clip, trim=2.5cm 0.2cm 12.8cm 0.3cm, width=1.0\columnwidth]{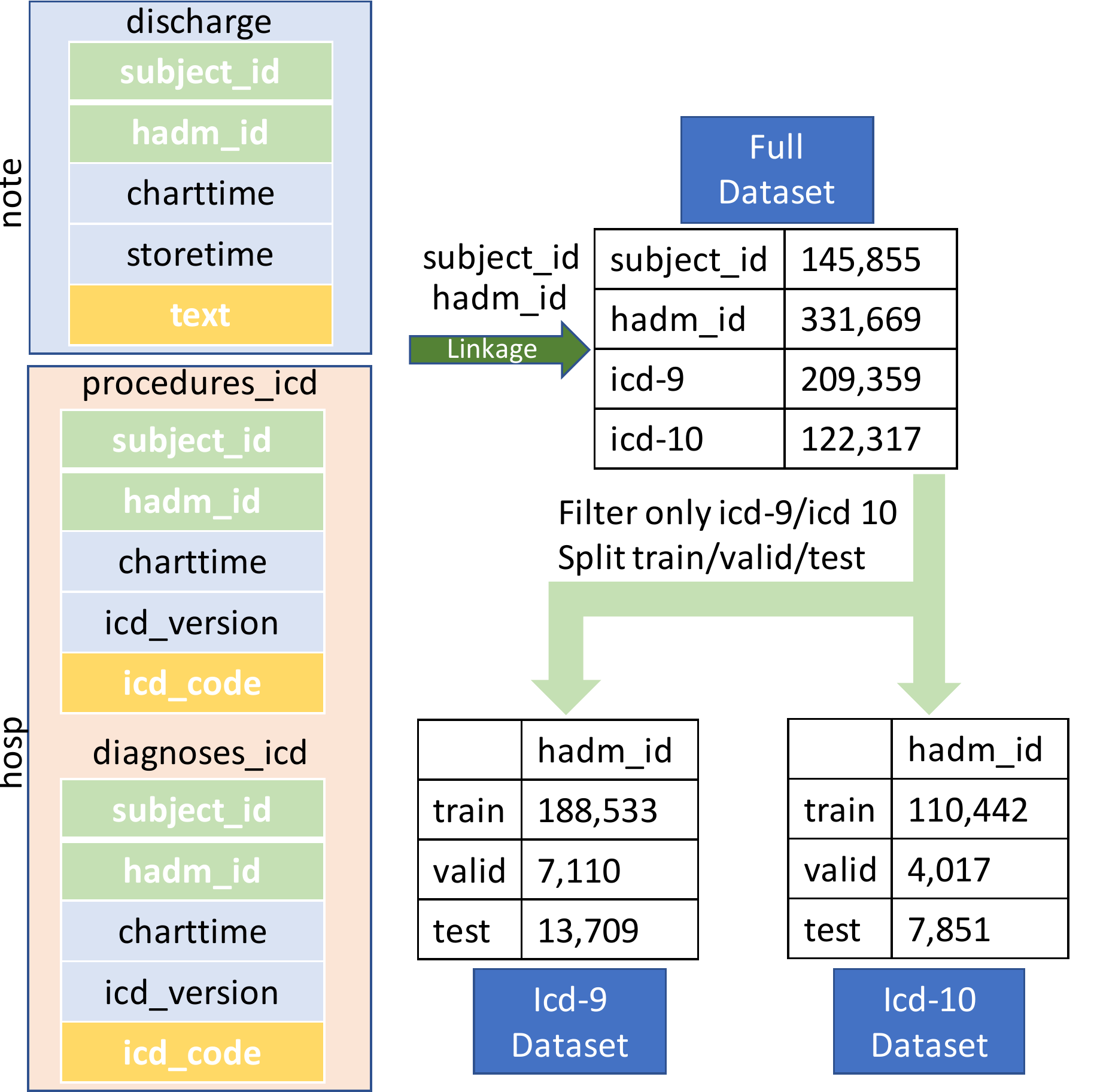}}
% \resizebox{1\columnwidth}{!}{%
% \includegraphics[width=0.5\textwidth]{imgs/MIMIC-IV-ICD_data_processing_workflow.png}
% }
% \vspace{-1em}
\caption{\small The workflow of data processing from raw data }
% \red{make it single column. we may need more space}}
    \label{fig:Data processing workflow}
\end{figure}
We standardized the terminology as follows: patients are identified by their \texttt{subject\_id}, and each patient may have multiple hospital admissions, denoted by hadm\_id. Both \texttt{subject\_id} and \texttt{hadm\_id} can be traced back to the MIMIC-IV database to follow a patient throughout their hospitalization. The data processing workflow, illustrated in Figure \ref{fig:Data processing workflow}, starts with the creation of a master dataset consisting of discharge notes and ICD codes. This is achieved by linking the 'discharge' table from the 'note' module to the 'procedures\_icd' and 'diagnoses\_id' tables in the 'hosp' module, using \texttt{subject\_id} and \texttt{hadm\_id} as primary keys. Generally, each discharge note corresponds to one pair of hospital admission id \texttt{(hadm\_id)} and patient id \texttt{(subject\_id)}. After filtering out discharge note pairs without ICD codes, the final master dataset comprises 331,669 hospital admissions for 145,855 patients.

Unlike MIMIC-III, which exclusively contains ICD-9 codes, MIMIC-IV encompasses both ICD-9 and ICD-10 codes. The master dataset includes 209,359 hospital admissions with ICD-9 codes and 122,317 hospital admissions with ICD-10 codes, compared to MIMIC-III's 52,726 documents. Additionally, there are seven admissions with both ICD code versions. In terms of unique labels, MIMIC-IV has 11,311 and 26,096 unique codes for ICD-9 and ICD-10 versions, respectively, while MIMIC-III has only 8,922 unique codes. Table \ref{tab: statistics-master-dataset} displays the basic statistics of the master dataset, juxtaposing them with MIMIC-III's statistics. Overall, MIMIC-IV features more documents and unique labels for both ICD-9 and ICD-10 code versions, although the number of tokens and labels is roughly equivalent. Evaluating existing methods from MIMIC-III in the MIMIC-IV context is advantageous for determining their performance in larger and more complex multilabel classification scenarios.

\begin{table*}[ht]
    \centering
    \small
    \begin{tabular}{lccccccc}
    \toprule
    & Train & Dev & Test & & Train & Dev & Test \\
    \multicolumn{4}{c}{MIMIC-III Full} & \multicolumn{2}{c}{MIMIC-III 50} \\
    \midrule
    \# Doc. & 47,723 & 1,631 & 3,372 & &8,066 &1,573 &1,729\\
    Avg \# words per Doc. & 1,434 & 1,724 & 1,731 & &1,478 & 1,739&1,731 \\
    Avg \# parent codes per Doc. & 13.7 & 15.4 & 15.9  & &5.3 &5.6 &5.7 \\ 
    Total \# parent codes & 1149 & 741 & 850  & & 39& 39&39\\
    Avg \# child codes per Doc. & 15.7 & 18.0 & 17.4  & & 5.7&5.9 &6.0\\ 
    Total \# unique child codes & 8,692 & 3,012 & 4,085  & &50 &50 & 50\\ 
    \midrule
    \multicolumn{4}{c}{MIMIC-IV-ICD9} & \multicolumn{2}{c}{MIMIC-IV-ICD9-50}\\
    \midrule
    \# Doc. & 188,533 & 7,110 & 13,709 & & 170,664 & 6,406 &12,405\\
    Average \# of words per Doc. & 1,459 & 1,472 & 1,460 & &1,499&1,516&1,501\\
    Avg \# of parent codes per Doc. & 12.1 & 12.2 & 12.0  & &4.6 & 4.7&4.7\\ 
    Total \# of unique parent codes & 1,230 & 954 & 1,041  & & 37& 37& 37\\
    Avg \# of child codes per Doc. & 13.4 & 13.5 & 13.3  & & 4.7& 4.8&4.8\\ 
    Total \# of unique child codes & 11,145 & 5,115 & 6,264  & &50 &50 &50\\

    \midrule
    \multicolumn{4}{c}{MIMIC-IV-ICD10} & \multicolumn{2}{c}{MIMIC-IV-ICD10-50}\\
    \midrule
    \# Doc. & 110,442 & 4,017 & 7,851 & &104,077& 3,805& 7,368\\
    Total \# of words per Doc. & 1,662 & 1,671 & 1,642 & &1,687 & 1,695& 1,669\\
    Avg \# parent codes per Doc. & 14.8 & 14.9 & 14.5 & &5.3&5.2&5.1\\ 
    Total \# of unique parent codes & 2,220 & 1,449 & 1,627 & &38 &38 &38\\
    Avg \# child codes per Doc. & 16.1 & 16.2 & 15.8 & &5.4 &5.4 &5.3\\ 
    Total \# unique child codes & 25,230 & 6,738 & 9,159 & & 50 &50 &50\\
    
    \bottomrule
    
    \end{tabular}
    \caption{Statistics of MIMIC-III and MIMIC-IV datasets under ICD-9 and ICD-10 codes settings.}
    \label{tab:statistic dataset}
\end{table*}

After obtaining the master dataset, it is necessary to divide it into the train, validation, and test sets. We followed the splitting procedures employed in \cite{mullenbach-etal-2018-explainable}. First, the master dataset is split to ensure that the patient or \texttt{subject\_id} for each train, validation, and test set does not overlap. Second, the dataset is partitioned based on the patient's percentage: 90\%, 3.33\%, and 6.67\% for training, development, and testing, respectively. This results in 188,533, 7,110, and 13,709 documents for the train, validation, and test sets of ICD-9 codes, and 110,442, 4,017, and 7,851 documents for their ICD-10 counterparts. To enable fair comparisons, we disclose our split by sharing the hadm\_id for each partition. %Table \ref{tab:statistic dataset} presents the statistics of each dataset and compares them with MIMIC-III.

\subsection{ICD Code Processing}

In the previous section, we discussed the full-label settings. In this section, we first construct datasets consisting only of the 50 most frequent labels, similar to previous work with MIMIC-III. Under this setting, we filter the train, validation, and test sets from the previous sections down to instances containing at least one of the top 50 most frequent codes. We remove instances containing none of the top 50 codes and the codes outside the top 50 codes. This results in 170,664 training, 6,406 validation, and 12,405 testing summaries for ICD-9 datasets, and 104,077 training, 3,805 validation, and 7,368 testing summaries for ICD-10 datasets.

In addition to creating the top 50 datasets, we also process the codes for studies that employ code ontology \citep{ijcai2020-461-vu} or descriptions for ICD code prediction \citep{yuan-etal-2022-code}. For ICD-9 diagnosis codes starting with 'E', which corresponds to external causes of injury, we represent the parent code using the first four characters of the code. For other diagnosis codes, we represent the parent codes with the first three characters. For example, diagnosis code E801.3 (Railway accident involving collision with other object and injuring pedal cyclist) belongs to the E801 category (Railway accident involving collision with other object), and diagnosis code 339.2 (Post-traumatic headache) has its parent code as 339 (Other headache syndromes). For ICD-9 procedure codes, we use the first two characters as the parent codes, e.g., procedure code 08.01 (Incision of lid margin) belongs to category 08 (Operations on eyelids). For ICD-10, we use the first three characters as parent codes; for instance, code Z00.01 (Encounter for general adult medical examination with abnormal findings) has its parent code as Z00 (Encounter for general examination without complaint, suspected, or reported diagnosis). We obtain code hierarchy and descriptions from the official 2023 ICD10 release by the Centers for Medicare and Medicaid Services\footnote{https://www.cms.gov/medicare/icd-10/2023-icd-10-cm}. ICD Code synonyms required for the MSMN model \cite{yuan-etal-2022-code} are obtained from UMLS \cite{umls2004}.
\subsection{Corpus}
The statistics of the resulting dataset are described in Table~\ref{tab:statistic dataset}. The table shows that the new ICD9 and ICD10 datasets contain more than thrice and twice the number of examples compared to MIMIC-III, respectively. As such, the number of unique ICD codes observed in the dataset increases. This is especially true for the transition to ICD10, which features a richer and more specific hierarchy of diagnoses and procedures and thus more billable codes to select from. Similar to MIMIC-III, these codes follow a natural long-tail distribution, where few codes appear often and the overwhelming majority is rare. Therefore, 50\% of the ICD10 codes label at most three discharge summaries (twelve for ICD9). Furthermore, 2.0\% and 6.3\% ICD10 codes appear only in the respective test set, which requires \emph{zero-shot learning} approaches to correctly predict these.

Given the domain, the vocabulary of the discharge summary documents is expectedly similar to MIMIC3, with 42\% and 40\% of tokens\footnote{after white-space tokenization, lower-casing stop-word and digit-only token removal} in ICD9 and ICD10 splits appearing in the vocabulary of MIMIC3, respectively. Vocabulary differs less for frequent terms, with the overlap for the top 100 terms being 72\% for ICD9 and 64\% for ICD10 splits.
% \alert{Viktor Input}
\begin{table*}[!t]
\centering
\resizebox{2\columnwidth}{!}{%
\begin{tabular}{lrrrrrrrrrrrr}\toprule
& \multicolumn{6}{c}{\bfseries MIMIC-IV-ICD9-Full} &\multicolumn{6}{c}{\bfseries MIMIC-IV-ICD9-50} \\
\multirow{2}{*}{Model} &\multicolumn{2}{c}{AUC} &\multicolumn{2}{c}{F1} & Precision & &\multicolumn{2}{c}{AUC} &\multicolumn{2}{c}{F1} &Precision  \\\cmidrule{2-6}\cmidrule{8-12}
&Macro &Micro &Macro &Micro &P@8 & &Macro &Micro &Macro &Micro &P@5 & \\\midrule
%\multicolumn{6}{c}
{\textbf{Single models}} \vspace{0.2em}\\
CAML \citep{mullenbach-etal-2018-explainable}& 93.45 & 99.29 & 11.06 & 57.28 & 64.91 & &93.07 &94.05 &65.33 &69.23 &58.64 \\
% MultiResCNN \citep{li2020multirescnn} &91.0 &98.6 &9.0 &55.2 &73.4 & &89.30 &92.04 &59.29 &66.24 &61.56 \\
% MSATT-KG \citep{10.1145/3357384.3357897}&91.0 &98.6 &8.5 &55.3 &72.8 & &91.40 &93.60 &63.80 &68.40 &64.40  \\
LAAT \citep{ijcai2020-461-vu}&95.18 &99.47 &13.12 &60.31 &67.47 & &94.88 &96.29 &69.99 &74.46& 62.01 \\
Joint LAAT \citep{ijcai2020-461-vu}&95.57 &99.49 &14.17 &60.37 &67.46 & & 94.92 & 96.31 &69.93 &74.33 &61.95 \\
% Our Model &95.26 &99.46 &14.00 &60.33 &67.40 & &94.89 &96.28 &70.97 &74.89 &62.00 \\
\midrule
\multicolumn{12}{c}
{\textbf{Models with External Data/Knowledge}} \vspace{0.2em}\\
MSMN \citep{yuan-etal-2022-code}& 96.79 &99.56 & 13.94 & 61.15 & 68.89 & & 95.13 & 96.46 & 71.85 & 75.78 & 62.60  \\
PLM-ICD \citep{huang-etal-2022-plm} & 96.61 & 99.53 & 14.40 & 62.45 & 70.34 &  &94.97 & 96.41 &71.35 & 75.46 &62.44\\
% KEPTLongformer \citep{Yang2022KnowledgeIP}&- &- &11.8 &59.9& 77.1 & &92.63 &94.76 &68.91 &72.85& 67.26\\
\bottomrule
\end{tabular}
}
\caption{Results of ICD9 code prediction models on the MIMIC-IV-ICD9-Full and MIMIC-IV-ICD9-50 test sets. 
% All our experiments are run five different random seeds and we report the mean results. The results of other models, except PLM-ICD, are collected from \citet{Yang2022KnowledgeIP}. For PLM-ICD, we follow the authors' instructions to reproduce the results.
}\label{tab:result_full_icd9}
\end{table*}
\section{Empirical Study}
\subsection{Baseline Methods}
Baseline Models. In this section, we present the baseline models utilized for comparison in our study:

\paragraph{CAML} The Convolutional Attention network for MultiLabel classification (CAML) was introduced by \citet{mullenbach-etal-2018-explainable}. It comprises a single-layer Convolutional Neural Network (CNN) and an attention layer that generates label-dependent representations for each ICD code.

\paragraph{LAAT} The Label Attention Model, proposed by \citet{ijcai2020-461-vu}, consists of a single bidirectional Long Short-Term Memory (LSTM) network that produces latent representations for clinical notes. The label attention layer applies a structured self-attention mechanism to generate label-specific document representations.

\paragraph{JointLAAT} The Hierarchical Joint Learning model of LAAT predicts the first level of ICD codes (parent labels) and uses them as additional input for the final label attention prediction. This approach helps address imbalanced and long-tail labels, training the model by minimizing the joint losses of both parent and child labels.

\paragraph{MSMN} The Multiple Synonyms Matching Network \citep{yuan-etal-2022-code} leverages synonyms for improved code representation learning through a multi-head-synonym attention and pooling mechanism.

\paragraph{PLM-ICD} The ICD Coding with Pretrained Language Models, proposed by \citet{huang-etal-2022-plm}, is a framework that employs pretrained language models to encode documents and uses the label attention layer from \citet{ijcai2020-461-vu} to enhance ICD coding prediction.

\subsection{Implementation and Evaluation}
To establish benchmarks for our ICD coding datasets, we executed the baseline models from the previous section using most of the hyperparameters reported in their respective papers. For clinical note preprocessing, we employed the standard regular expression tokenizer from the Natural Language Toolkit (NLTK) to tokenize the text into a list of word characters, convert the text to lowercase, and truncate it to the maximum length for each model. For training, we primarily adjusted the batch size to accommodate our GPUs, as MIMIC-IV datasets are larger and contain more labels than MIMIC-III. CAML, LAAT, and JointLAAT were trained using a single 16GB Tesla P100 GPU. Meanwhile MSMN, unlike for MIMIC-III, required more than 32 GB of memory and was thus trained on an 80GB A100 GPU. PLM-ICD was optimised using two 16 GB V100 GPUs.

As with MIMIC-III, we employed macro and micro AUC and F1, as well as precision@k (with $k=8$ for full settings and $k=5$ for top-50 settings). Micro scores average the performance across all label-instance pairs, providing an aggregate measure of performance across all labels. This is particularly useful for imbalanced datasets, as it assigns equal weight to each instance, regardless of label distribution. On the other hand, macro-F1 calculates an unweighted average of F1 scores for each label, treating all labels as equally important. This is especially useful for datasets with imbalanced label distribution, as it ensures that the performance on rare labels is not overshadowed by the performance on more frequent labels. Since MIMIC-IV has a similar average number of codes per document as MIMIC-III, precision@k remains appropriate.

\subsection{Benchmark Results}

\begin{table*}[!t]
\centering
\resizebox{2\columnwidth}{!}{%
\begin{tabular}{lrrrrrrrrrrrr}\toprule
& \multicolumn{6}{c}{\bfseries MIMIC-IV-ICD10-Full} &\multicolumn{6}{c}{\bfseries MIMIC-IV-ICD10-50} \\
\multirow{2}{*}{Model} &\multicolumn{2}{c}{AUC} &\multicolumn{2}{c}{F1} & Precision & &\multicolumn{2}{c}{AUC} &\multicolumn{2}{c}{F1} &Precision  \\\cmidrule{2-6}\cmidrule{8-12}
&Macro &Micro &Macro &Micro &P@8 & &Macro &Micro &Macro &Micro &P@5 & \\\midrule
%\multicolumn{6}{c}
{\textbf{Single models}} \vspace{0.2em}\\
CAML \citep{mullenbach-etal-2018-explainable}& 89.91 & 98.79 & 4.07 & 52.67 & 64.43 & &91.05 &93.18 &64.30 &67.56 &59.58 \\
% MultiResCNN \citep{li2020multirescnn} &91.0 &98.6 &9.0 &55.2 &73.4 & &89.30 &92.04 &59.29 &66.24 &61.56 \\
% MSATT-KG \citep{10.1145/3357384.3357897}&91.0 &98.6 &8.5 &55.3 &72.8 & &91.40 &93.60 &63.80 &68.40 &64.40  \\
LAAT \citep{ijcai2020-461-vu}&92.96 &99.14 &4.47 &55.40 &66.97 & &93.21 &95.49 &68.15 &72.56 &64.39 \\
Joint LAAT \citep{ijcai2020-461-vu}&93.64 &99.27 &5.71 &55.89 &66.89 & & 93.39 & 95.57 &68.41 &72.85 &64.49 \\
% Our Model &96.56 &99.47 &6.03 &56.46 &67.34 & &93.37 &95.58 &69.01 &73.17 &64.40 \\
\midrule
\multicolumn{12}{c}
{\textbf{Models with External Data/Knowledge}} \vspace{0.2em}\\
MSMN \citep{yuan-etal-2022-code}& 97.07 & 99.61 & 5.42 & 55.91 & 67.66 & & 93.60 & 95.69 & 70.31 & 74.15 & 65.16  \\
PLM-ICD \citep{huang-etal-2022-plm} & 91.85 & 99.02 & 4.90 & 56.95 & 69.47 & & 93.37 & 95.61 & 69.01 & 73.27 & 64.57 \\
% KEPTLongformer \citep{Yang2022KnowledgeIP}&- &- &11.8 &59.9& 77.1 & &92.63 &94.76 &68.91 &72.85& 67.26\\
\bottomrule
\end{tabular}
}
\caption{Results of ICD10 code prediction models on MIMIC-IV-ICD10-Full and MIMIC-IV-ICD10-50 test sets. 
% All our experiments are run five different random seeds and we report the mean results. The results of other models, except PLM-ICD, are collected from \citet{Yang2022KnowledgeIP}. For PLM-ICD, we follow the authors' instructions to reproduce the results.
}\label{tab:result_full_icd10}
\end{table*}

\paragraph{MIMIC-IV-ICD-9} The results presented in Table \ref{tab:result_full_icd9} indicate that, among the "single" models that do not rely on external data or knowledge, JointLAAT achieves the highest performance for MIMIC-IV-ICD-9-Full settings with a macro AUC of $95.57\%$, micro AUC of $99.49\%$, macro F1 of $14.17\%$, micro F1 of $60.37\%$, and precision@8 of $67.46\%$. On the other hand, LAAT also attains mostly the highest results in MIMIC-IV-ICD-9-50 settings, with macro AUC of $94.88\%$, micro AUC of $96.29\%$, macro F1 of $69.99\%$, macro F1 of $74.46\%$, and precision@5 of $62.01\%$. Among the models utilizing external data or knowledge, PLM-ICD achieves the best performance under MIMIC-IV-ICD-9-Full settings, with macro AUC of $96.61\%$, micro AUC of $99.53\%$, macro F1 of $14.40\%$, micro F1 of $62.45\%$, and precision@8 of $70.34\%$. Meanwhile, MSMN attains the best results under MIMIC-IV-ICD-9-50 settings, with macro AUC of $95.13\%$, micro AUC of $96.46\%$, micro F1 of $71.85\%$, macro F1 of $75.78\%$, and precision@5 of $62.60\%$.

\paragraph{MIMIC-IV-ICD-10} The results in Table \ref{tab:result_full_icd10} indicate that under MIMIC-IV-ICD-10-Full settings, LAAT achieves the highest performance among "single" models that do not rely on external data or knowledge, with macro AUC, micro AUC, micro F1, macro F1, and precision@8 of $93.64\%$, $99.27\%$, $5.71\%$, $55.89\%$ and  $66.89\%$, respectively. Precisely, it also attains $93.39\%$, $95.57\%$, $68.41\%$, $72.85\%$, and $64.49\%$ in macro AUC, micro AUC, micro F1, macro F1, and precision@5, respectively, which are mostly the highest in the same setting for MIMIC-IV-ICD-10-50. In models utilizing external data or knowledge, PLM-ICD achieves the best results under MIMIC-IV-ICD-10-Full settings, with macro AUC, micro AUC, micro F1, macro F1, and precision@8 of $91.85\%$, $99.02 \%$, $4.90\%$, $56.95\% $ and $69.47\%$. Meanwhile, MSMN achieves the best results under MIMIC-IV-ICD-9-50 settings, with macro AUC, micro AUC, micro F1, macro F1, and precision@5 of $93.60\%$, $95.69\%$, $70.31\%$, $74.15\%$, and $65.16\%$. These results demonstrate the varied performance of models under different settings, emphasizing the importance of model selection based on the specific context and requirements.

\subsection{Ablation study}
To assess the effectiveness of our baseline models, we perform an ablation study on both the MIMIC-IV-ICD-9-Full and MIMIC-IV-ICD-10-Full datasets, comparing the two best-performing models: Multiple Synonyms Matching Network (MSMN) and ICD Coding with Pretrained Language Models (PLM-ICD), which are the top models in our benchmark. We examine the performance of these models on labels grouped by their frequency of appearance. To gain deeper insights into the models' predictions, we categorize medical codes into five groups based on their frequencies in MIMIC-IV-Full-ICD-9 and MIMIC-IV-Full-ICD-10 datasets: $1-10, 11-50, 51-100, 101-500, >500$. The statistics of all groups in both datasets are presented in Table \ref{tab: label frequency distribution Mimic-IV-ICD}.
\begin{table}[!t]
\centering
\small
\begin{tabular}{ccc}
\toprule
Frequency range & No. of ICD-9 codes  & No. of ICD-10 codes\\
\midrule
1-10 &5,262 & 18,483 \\
11-50 &2,706 & 4,471 \\
51-100 & 911 & 1,179\\
101-500 & 1,492& 1,337 \\
>500 & 853 & 626\\
\bottomrule
\end{tabular}
\caption{Label Frequency Distribution of Mimic-IV-ICD-9-Full and MIMIC-IV-ICD-10-Full}\label{tab: label frequency distribution Mimic-IV-ICD}
\end{table}
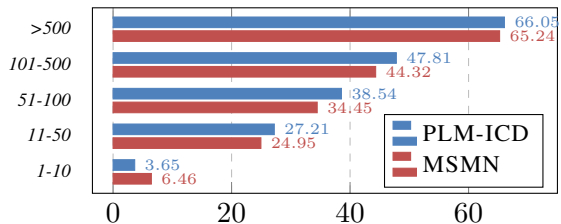
\begin{figure}[!t]
    \begin{tikzpicture}
    \begin{axis}[
        xbar=-10pt,
        y tick label style={anchor=north east, align=right,text width=1cm, font=\scriptsize\itshape},
        legend style={font=\small},
        height=10.5em,
        major y tick style = transparent,
        % ylabel=\emph{Label Frequency},
        % xlabel=\emph{Micro F1},
        width  = \columnwidth,
        bar width=4pt,
        xmajorgrids=true,
        %y tick label as interval,
        %ymajorgrids=true,
        x grid style=dashed,
        legend pos=south east,
        legend cell align={left},
        legend columns=1,
        ytick={1,2,3,4,5},
        ymin=-0.1, ymax=5.1,
        xmax=75,
        nodes near coords,
        every node near coord/.append style={font=\tiny},
        yticklabels={1-10, 11-50, 51-100, 101-500, >500},
    ]
    \addplot[legend entry=\textsc{PLM-ICD}, color=bblue, fill=bblue]  coordinates {(3.65,0.47) (27.21,1.47) 
    (38.54, 2.47) (47.81,3.47) (66.05,4.47)};
     \addplot+[legend entry=\textsc{MSMN}, color=rred, fill=rred]  coordinates {(6.46,0.53) (24.95,1.53) (34.45, 2.53) (44.32,3.53) (65.24,4.53)};
    \end{axis}
    \end{tikzpicture}
    %\vspace{2\baselineskip}
    
    \caption{Comparison of Micro-F1 scores between PLM-ICD and MSMN on labels with different Mimic-IV-ICD-9-Full test set frequencies.}
    \label{fig:Micro F1 by label frequency group Mimic-IV-ICD-9}
\end{figure}

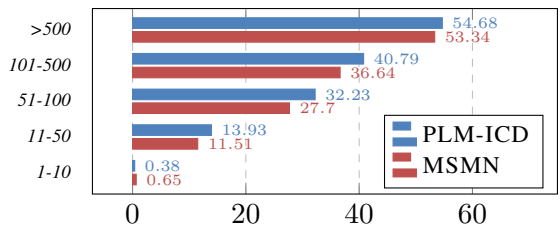
\begin{figure}[!t]
    \begin{tikzpicture}
    \begin{axis}[
        xbar=-10pt,
        y tick label style={anchor=north east, align=right,text width=1cm, font=\scriptsize\itshape},
        legend style={font=\small},
        height=10.5em,
        major y tick style = transparent,
        % ylabel=\emph{Label Frequency},
        % xlabel=\emph{Micro F1},
        width  = \columnwidth,
        bar width=4pt,
        xmajorgrids=true,
        %y tick label as interval,
        %ymajorgrids=true,
        x grid style=dashed,
        legend pos=south east,
        legend cell align={left},
        legend columns=1,
        ytick={1,2,3,4,5},
        ymin=-0.1, ymax=5.1,
        xmax=75,
        nodes near coords,
        every node near coord/.append style={font=\tiny},
        yticklabels={1-10, 11-50, 51-100, 101-500, >500},
    ]
    \addplot[legend entry=\textsc{PLM-ICD}, color=bblue, fill=bblue]  coordinates {(0.38,0.47) (13.93,1.47) (32.23,2.47) (40.79,3.47) (54.68,4.47)};
     \addplot+[legend entry=\textsc{MSMN}, color=rred, fill=rred]  coordinates {(0.65,0.53) (11.51,1.53) (27.70, 2.53) (36.64,3.53) (53.34,4.53)};
    \end{axis}
    \end{tikzpicture}
    %\vspace{2\baselineskip}
    
    \caption{Comparison of Macro-F1 scores between PLM-ICD and MSMN on labels with different Mimic-IV-ICD-9-Full test set frequencies.}
    \label{fig:Macro F1 by label frequency group Mimic-IV-ICD-9}
\end{figure}
\paragraph{MIMIC-IV-ICD-9}
We compare the Micro F1 and Macro F1 scores across different groups in Figures~\ref{fig:Micro F1 by label frequency group Mimic-IV-ICD-9} and \ref{fig:Macro F1 by label frequency group Mimic-IV-ICD-9}. In general, PLM-ICD outperforms MSMN in most groups. For Micro-F1, the differences are particularly noticeable in the frequent groups: $1\%$ in the $>500$ group versus $3\%$ in the $101-500$, $51-100$, and $11-50$ groups, while PLM-ICD performs worse than MSMN in the $1-10$ group, which contains the majority of codes. For Macro-F1, the differences are similar in the frequent groups: $1\%$ in the $>500$ group, $4\%$ in the $101-500$ group, $5\%$ in the $51-100$ group, and $2\%$ in the $11-50$ group, while PLM-ICD still performs slightly worse than MSMN in the $1-10$ group. It appears that in the ICD-9 setting, PLM-ICD learns better than MSMN.

% \begin{table}[!t]
% \centering
% \small
% \begin{tabular}{rr}\toprule
% Frequency range & Number of codes  \\
% \midrule
% 1-10 &18,483 \\
% 11-50 &4,471 \\
% 51-100 & 1,179 \\
% 101-500 & 1,337 \\
% >500 & 626 \\

% \bottomrule
% \end{tabular}
% \caption{Label Frequency Distribution of Mimic-IV-ICD-10-Full}\label{tab: label frequency distribution Mimic-IV-ICD-10}
% \end{table}
\begin{figure}[!t]
    \begin{tikzpicture}
    \begin{axis}[
        xbar=-10pt,
        y tick label style={anchor=north east, align=right,text width=1cm, font=\scriptsize\itshape},
        legend style={font=\small},
        height=10.5em,
        major y tick style = transparent,
        % ylabel=\emph{Label Frequency},
        % xlabel=\emph{Micro F1},
        width  = \columnwidth,
        bar width=4pt,
        xmajorgrids=true,
        %y tick label as interval,
        %ymajorgrids=true,
        x grid style=dashed,
        legend pos=south east,
        legend cell align={left},
        legend columns=1,
        ytick={1,2,3,4,5},
        ymin=-0.1, ymax=5.1,
        xmax=75,
        nodes near coords,
        every node near coord/.append style={font=\tiny},
        yticklabels={1-10, 11-50, 51-100, 101-500, >500},
    ]
    \addplot[legend entry=\textsc{PLM-ICD}, color=bblue, fill=bblue]  coordinates {(0.49,0.47) (12.69,1.47) (28.73,2.47) (43.95,3.47) (64.10,4.47)};
     \addplot+[legend entry=\textsc{MSMN}, color=rred, fill=rred]  coordinates {(3.16,0.53) (17.62,1.53) (29.73, 2.53) (42.26,3.53) (62.87,4.53)};
    \end{axis}
    \end{tikzpicture}
    %\vspace{2\baselineskip}
    
    \caption{Comparison of Micro-F1 scores between PLM-ICD and MSMN on labels with different Mimic-IV-ICD-10-Full test set frequencies.}
    \label{fig:Micro F1 by label frequency group Mimic-IV-ICD-10}
\end{figure}
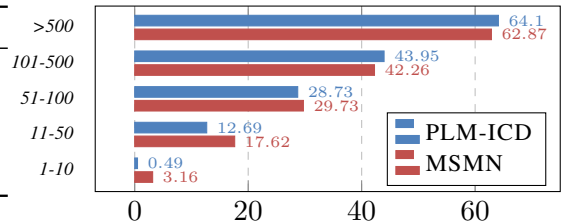

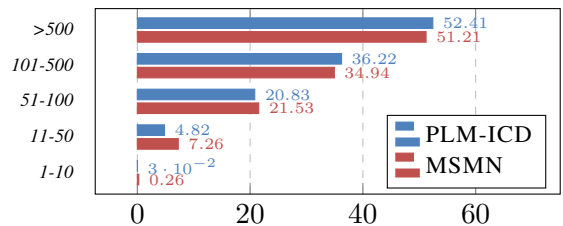
\begin{figure}[!t]
    \begin{tikzpicture}
    \begin{axis}[
        xbar=-10pt,
        y tick label style={anchor=north east, align=right,text width=1cm, font=\scriptsize\itshape},
        legend style={font=\small},
        height=10.5em,
        major y tick style = transparent,
        % ylabel=\emph{Label Frequency},
        % xlabel=\emph{Micro F1},
        width  = \columnwidth,
        bar width=4pt,
        xmajorgrids=true,
        %y tick label as interval,
        %ymajorgrids=true,
        x grid style=dashed,
        legend pos=south east,
        legend cell align={left},
        legend columns=1,
        ytick={1,2,3,4,5},
        ymin=-0.1, ymax=5.1,
        xmax=75,
        nodes near coords,
        every node near coord/.append style={font=\tiny},
        yticklabels={1-10, 11-50, 51-100, 101-500, >500},
    ]
    \addplot[legend entry=\textsc{PLM-ICD}, color=bblue, fill=bblue]  coordinates {(0.03,0.47) (4.82,1.47) (20.83,2.47) (36.22,3.47) (52.41,4.47)};
     \addplot+[legend entry=\textsc{MSMN}, color=rred, fill=rred]  coordinates {(0.26,0.53) (7.26,1.53) (21.53, 2.53) (34.944,3.53) (51.21,4.53)};
    \end{axis}
    \end{tikzpicture}
    %\vspace{2\baselineskip}
    
    \caption{Comparison of Macro-F1 scores between PLM-ICD and MSMN on labels with different Mimic-IV-ICD-10-Full test set frequencies.}
    \label{fig:Macro F1 by label frequency group Mimic-IV-ICD-10}
\end{figure}
\paragraph{MIMIC-IV-ICD-10}
We compare the Micro F1 and Macro F1 scores across different groups in Figures~\ref{fig:Micro F1 by label frequency group Mimic-IV-ICD-10} and \ref{fig:Macro F1 by label frequency group Mimic-IV-ICD-10}. In general, PLM-ICD outperforms MSMN in most groups. The differences in micro F1 are particularly noticeable in the more frequent groups ($2\%$ in $>500$ and $101-500$ groups versus), while PLM-ICD performs worse than MSMN in the $1-10$, $11-50$, and $>500$ groups, which contains the majority of codes. We observe the similar pattern in macro F1: the differences are $1\%$) in groups $>500$, $101-500$, while PLM-ICD performs worse than MSMN in the $51-100$ and $11-50$ group but slightly better in the $1-10$ group which contains the majority of codes.One possible explanation for this is that both models can learn from a few examples in very rare codes; however, with the assistance of multiple synonyms, MSMN can better match the semantic meaning of the codes to the medical notes compared to PLM-ICD, which does not consider code descriptions and relies solely on code embeddings. In the more frequent groups, PLM-ICD outperforms MSMN due to its superior encoder from large pretrained models. This suggests a potential future direction for improving both code representation and medical note representation using large language models.

\section{Related Work}

Many ICD coding datasets exist in various languages and for various medical stages, such as ICD coding for admission notes, progress notes, or discharge notes. However, not all of them are publicly available due to the privacy concerns associated with clinical data. The most popular datasets are the MIMIC databases. MIMIC-III (Medical Information Mart for Intensive Care III) was one of the first large, freely-available databases consisting of deidentified health-related data for patients admitted to critical care units at the Beth Israel Deaconess Medical Center from 2001 to 2012. The database includes information such as demographics, bedside vital sign measurements, laboratory test results, procedures, medications, caregiver notes, imaging reports, and mortality (both in and out of the hospital). MIMIC-III supports a wide range of analytic studies, including epidemiology, clinical decision-rule improvement, and electronic tool development. \citep{mullenbach-etal-2018-explainable, shi-et-al-2027-automated-icd} are two of the first studies to publish a data pipeline for processing discharge summaries and matching them with ICD-9 codes, forming the MIMIC-III-full and MIMIC-III-top-50 sets, which became the benchmark for MIMIC-III ICD coding.

MIMIC-IV is the latest database containing real hospital stays for patients admitted to a tertiary academic medical center in Boston, MA, USA. It contains comprehensive information about each patient during their hospital stay, such as laboratory measurements, medications administered, and documented vital signs. The database aims to support a wide variety of research in healthcare. MIMIC-IV builds upon the success of MIMIC-III and incorporates numerous improvements. Several benchmarks and pipelines have been developed for MIMIC-IV to utilize its extensive dataset for various medical tasks: \citep{DBLP:conf/ml4h/GuptaGCDPB22} propose a data processing pipeline for extracting, cleaning, and preprocessing MIMIC-IV data for time-series tasks such as mortality prediction and readmission admission, while \citep{DBLP:conf/amia/XieZLTLRCCWDOGL22} propose a benchmark for emergency department (ED) triage, critical outcome prediction, and reattendance prediction at ED triage. However, there is no benchmark for ICD coding for MIMIC-IV. Our work aims to provide a processing procedure for this task, allowing researchers to process data, reproduce results, and conduct further research on top of it.

\section{Conclusion and Recommendations}

The field of machine learning is witnessing a surge in research focused on building clinical predictive models using EHR data. These models effectively capture the complexities in EHR data and aid in predicting future outcomes. MIMIC datasets encourage research in this domain by providing a unique and extensive EHR dataset for researchers to explore. In this study, we establish a standardized benchmark for ICD coding on MIMIC-IV, covering both ICD-9 and ICD-10 codes. This process involves converting raw data into a task-specific format and applying popular deep learning baseline methods to the new datasets. Following the example set by \citep{mullenbach-etal-2018-explainable}, we make our data processing code open-source, enabling researchers to reproduce and enhance the results. 

In the future, we plan to expand our benchmark by adding more baselines, potentially incorporating relevant features such as drug codes and patient vitals. Since ICD codes play a crucial role in enhancing patient care, facilitating research, and ensuring accurate communication among healthcare providers, we aim to extend the use of clinical notes for joint prediction of ICD codes and readmission, triage, and mortality prediction tasks. By openly sharing our data processing code with the community, we hope to inspire others to join us in the ongoing refinement of this benchmark.

\bibliography{anthology,custom}
\bibliographystyle{acl_natbib}

\end{document}